\documentclass[11pt,a4paper]{article}
\pdfoutput=1

\usepackage[hyperref]{emnlp2018}
\usepackage{times}
\usepackage{latexsym}
\usepackage{url}

\aclfinalcopy 

\usepackage{amsmath}
\usepackage{booktabs}
\usepackage{pgfplots}
\usepackage{pgfplotstable}
\usepackage{booktabs}
\usepackage{array}
\usepackage{colortbl}
\usepackage{filecontents}
\usetikzlibrary{patterns}
\usetikzlibrary{calc,positioning}
\usepgfplotslibrary{statistics}
\usepgfplotslibrary{groupplots}
\usepackage{tikzsymbols}
\usepackage{multicol}
\usepackage{ulem}
\usepackage{microtype}
\usepackage{paralist}
\usepackage{enumitem}
\setlist{noitemsep}
\usepackage{breqn}
\usepackage{amssymb}
\usepackage{subcaption}

\definecolor{bblue}{HTML}{4F81BD}
\definecolor{rred}{HTML}{C0504D}
\definecolor{ggreen}{HTML}{9BBB59}
\definecolor{ppurple}{HTML}{9F4C7C}
\definecolor{oorange}{HTML}{F08000}

\title{Dual Conditional Cross-Entropy Filtering of Noisy Parallel Corpora}

\author{Marcin Junczys-Dowmunt \\
Microsoft \\
1 Microsoft Way\\
Redmond, WA 98121, USA}

\date{}

\begin{document}
\maketitle
\begin{abstract}
In this work we introduce dual conditional cross-entropy filtering for noisy parallel data. For each sentence pair of the noisy parallel corpus we compute cross-entropy scores according to two inverse translation models trained on clean data. We penalize divergent cross-entropies and weigh the penalty by the cross-entropy average of both models. Sorting or thresholding according to these scores results in better subsets of parallel data. We achieve higher BLEU scores with models trained on parallel data filtered only from Paracrawl than with models trained on clean WMT data. We further evaluate our method in the context of the WMT2018 shared task on parallel corpus filtering and achieve the overall highest ranking scores of the shared task, scoring top in three out of four subtasks.
\end{abstract}

\section{Introduction}

Recently, large web-crawled parallel corpora which are meant to rival non-public resources held by popular machine translation providers have been made publicly available to the research community in form of the Paracrawl corpus.\footnote{\url{https://paracrawl.eu}} At the same time, it has been shown that neural translation models are far more sensitive to noisy parallel training data than phrase-based statistical machine translation methods \cite{khayrallah-koehn:2018:WNMT2018,DBLP:journals/corr/abs-1711-02173}. This creates the need for data selection methods that can filter harmful sentence pairs from these large resources.

In this paper, we introduce dual conditional cross-entropy filtering, a simple but effective data selection method for noisy parallel corpora. We think of it as the missing adequacy component to the fluency aspects of cross-entropy difference filtering by \newcite{moore-lewis:2010:Short}.
Similar to Moore-Lewis filtering for monolingual data, we directly select samples that have the potential to improve perplexity (and in our case translation performance) of models trained with the filtered data.

This is different from \newcite{axelrod} who simply expand \citeauthor{moore-lewis:2010:Short} filtering to both sides of the parallel corpus. We use conditional probability distributions and enforce agreement between inverse translation directions. 

In most cases, neural translation models are trained to minimize perplexity (or cross-entropy) on a training set. Our selection criterion includes the optimization criterion of neural machine translation which we approximate by using neural translation models pre-trained on clean seed data. 

We evaluated our method in the context of the WMT2018 Shared Task on Parallel Corpus Filtering \cite{parallel-filtering-task:2018:WMT} and submitted our best method to the task. Although we only optimized for one of the four subtasks of the shared task, our submission scored highest for three out of four subtasks and third for the fourth subtask; there were 48 submissions to each subtask in total. 





\section{WMT 2018 shared task on parallel corpus filtering}


We quote the shared task description provided by the organizers on the task website\footnote{\url{http://www.statmt.org/wmt18/parallel-corpus-filtering.html}} and add citations where appropriate:
The organizers ``provide a very noisy 1 billion word (English token count) German-English corpus crawled from the web as part of the Paracrawl project'' and ``ask participants to subselect sentence pairs that amount to (a) 100 million words, and (b) 10 million words. The quality of the resulting subsets is determined by the quality of a statistical machine translation --- Moses, phrase-based \cite{conf/acl/KoehnHBCFBCSMZDBCH07} --- and a neural machine translation system --- Marian \cite{marian} --- trained on this data.''
The organizers note that the task is meant to address ``the challenge of data quality and not domain-relatedness of the data for a particular use case.'' 
They discourage participants from sub-sampling the corpus for relevance to the news domain and announce that more emphasis will be put on undisclosed test sets rather than the WMT2018 test set.

Furthermore the organizers remark that ``the provided raw parallel corpus is the outcome of a processing pipeline that aimed from high recall at the cost of precision, so it is very noisy. It exhibits noise of all kinds (wrong language in source and target, sentence pairs that are not translations of each other, bad language, incomplete or bad translations, etc.)'' 
It is allowed to use the 2018 news translation task data for German-English (without the Paracrawl parallel corpus) to train components of our methods.

\subsection{Sub-sampling based on submitted scores}
Participants submit files with numerical scores, one score per line of the original unfiltered parallel corpus. A tool provided by the organizers takes as input the scores and the German and English corpus halves in form of raw text. Higher scores are better. The tool first determines at which best thresholds 10M and 100M words can be collected and next creates two data sets containing all sentences with scores above the two selected respective thresholds. Systems trained on these data sets are used for evaluation by the organizers (4 systems per submission) and for development purposes by task participants. 

We focus on the 100M sub-task for neural machine translation systems as this is closest to our interests of finding as much relevant data as possible in large noisy parallel corpora. We only develop systems for this scenario.

\subsection{Neural machine translation evaluation}
\label{arch}

As required by the shared task, we use Marian \cite{marian} to train our development systems. 
%
We follow the recommended settings quite closely in terms of model architecture, but change training settings, favoring hyper-parameters that lead to quicker convergence during our own development phase.
We switched off synchronous ADAM in favor of asynchronous ADAM, increased the evaluation frequency to once per 5000 updates and increased work-space size to 5000MB per GPU. We also set the initial learning-rate to 0.0003 instead of 0.0001 and used an inverse square-root decaying scheme for the learning rate \cite{NIPS2017_7181} that started after 16,000 updates. We removed dropout of source and target words and decreased variational dropout from 0.2 to 0.1 \cite{gal2016theoretically}. With these settings, our models usually converged within 10 to 15 hours of training on four NVidia Titan Xp GPUs. Convergence was assumed if perplexity did not improve for 5 consecutive evaluation steps. We evaluated on the provided WMT2016 and WMT2017 test sets. 

\section{Scores and experiments}

We produce a single score $f(x,y)$ per sentence pair $(x,y)$ as the product of partial scores $f_i(x,y)$:
\begin{equation}
f(x,y) = \prod_{i} f_i(x,y).
\end{equation}

Partial scores take values between 0 and 1, as does the total score $f$. Partial scores that might generate values outside that range are clipped. We assume that sentence pairs with a score of 0 are excluded from the training data.\footnote{This is only guaranteed by the selection algorithm of the shared task if more than 100M words appear in sentence pairs scored with non-zero scores. However, we did not encounter situations where we got close or below that boundary.} 

In this section, we describe the scores explored in this work and present results on the development data.

\subsection{Experimental baselines}

Following the training recipe in Section \ref{arch}, we first trained a model (``WMT18-full'' in Table~\ref{dev}) on the admissible parallel WMT18 data for German-English (excluding Paracrawl). This model is only used for the computation of reference BLEU scores.

Next, we trained a German-English model on randomly scored Paracrawl data only (``random'' in Table~\ref{dev}). The random scores -- uniformly sampled values between 0 and 1 -- were used to select representative data consisting of 100M words from unprocessed Paracrawl while using the threshold-based selection tool provided by the shared task organizers. Results for WMT16 and WMT17 test sets for both systems are shown in Table~\ref{dev}. The Paracrawl-trained systems (random) has dramatically worse BLEU scores than the WMT18-trained system. Upon manual inspection, we see many untranslated and partially copied sentences in the case of the randomly-selected Paracrawl system. 

\begin{table}[t]
\centering
\begin{tabular}{lp{5.5cm}}\toprule
Model & Description \\ \midrule
$W_{\mathrm{en}}$ & RNN language model trained on 1M sentences from English WMT monolingual news data 2015-2017 \\ 
$P_{\mathrm{en}}$ & RNN language model trained on 1M sentences from target (English) side of Paracrawl \\ \midrule
$W_{\mathrm{de}\rightarrow\mathrm{en}}$ & German-English translation model trained on WMT parallel data \\ $W_{\mathrm{en}\rightarrow\mathrm{de}}$ & English-German translation model trained on WMT parallel data \\ 
\midrule
$W_{\mathrm{de}\leftrightarrow\mathrm{en}}$ & Translation model trained on union of German-English and English-German WMT parallel data\\
\bottomrule
\end{tabular}
\caption{Helper models trained for various scorers. All models are neural models, we do not use n-gram or phrase-based models. WMT parallel data excludes Paracrawl data.}\label{helper}
\end{table}

\subsection{Language identification}

We noticed that the provided sentence pairs do not seem to have been subjected to language identification and simply used the Python \texttt{langid} package to assign a language code to each sentence in a sentence pair. We did not restrict the inventory of languages beforehand as we wanted the tool to propose a language if that language wins against all other candidates. We only accepted sentence pairs where both elements of a pair had been assigned the desired languages (German for source, English for target). The result is our first non-trivial score:

\begin{equation*}
\mathrm{lang}(x, l) = \left\{ 
\begin{array}{ll}
1 & \mathrm{if \; \textsc{LangID}}(x) = l \\ 
0 & \mathrm{otherwise} \\
\end{array} \right.
\end{equation*}
\begin{equation}
\mathrm{de\textrm{-}en}(x,y) = \mathrm{lang}(x, \textrm{``de''}) \cdot \mathrm{lang}(y, \textrm{``en''}) 
\end{equation}

This is a very harsh but also very effective filter that removes nearly 70\% of the parallel sentence candidates. As a beneficial side-effect of language identification many language-ambiguous fragments which contain only little textual information are discarded, e.g.~sentences with lots of numbers, punctuation marks or other non-letter characters. The identification tool gets confused by the non-textual content and selects a random language. 

We combined the $\mathrm{de\textrm{-}en(x,y)}$ filter with the random scores and trained a corresponding system (de-en$\cdot$random). As we see in Table \ref{dev}, this strongly improved the results on both dev sets. When reviewing the translated development sets, we did not see any copied/untranslated sentences in the output.  

\subsection{Dual conditional cross-entropy filtering}
\label{dual}

The scoring method introduced in this section is our main contribution. While inspired by cross-entropy difference filtering for monolingual data \cite{moore-lewis:2010:Short}, our method does not aim for monolingual domain-selection effects. Instead we try to model a bilingual adequacy score. 

\citeauthor{moore-lewis:2010:Short} (see next section) quantify the directed disagreement (signed difference) of similar distributions (two language models over the same language) trained on dissimilar data (different monolingual corpora). A stronger degree of separation between the two models indicates more interesting data.

In contrast, we try to find maximal symmetric agreement (minimal absolute difference) of dissimilar distributions (two translation models over inverse translation directions) trained on the same data (same parallel corpus). Concretely, for a sentence pair $(x,y)$ we calculate a score:
\begin{equation}
\begin{aligned}
  & \left| H_{A}(y|x) - H_{B}(x|y) \right| \\
+ & \frac{1}{2} \left( H_{A}(y|x) + H_{B}(x|y) \right) \label{dualscore}
\end{aligned}
\end{equation}
where $A$ and $B$ are translation models trained on the same data but in inverse directions, and $H_M(\cdot|\cdot)$ is the word-normalized conditional cross-entropy of the probability distribution $P_M(\cdot|\cdot)$ for a model $M$:
\begin{equation*}
\begin{aligned}
H_M(y|x) =& -\frac{1}{|y|}\log P_M(y|x) \\
 =& -\frac{1}{|y|} \sum_{t = 1}^{|y|}\log P_M(y_t|y_{<t},x).
\end{aligned}
\end{equation*}

The score (denoted as dual conditional cross-entropy) has two components with different functions: the absolute difference $\left| H_{A}(y|x) - H_{B}(x|y) \right|$ measures the agreement between the two conditional probability distributions, assuming that (word-normalized) translation probabilities of sentence pairs in both directions should be roughly equal. We want disagreement to be low, hence this value should be close to 0. 

However, a translation pair that is judged to be equally improbable by both models will also have a low disagreement score. Therefore we weight the agreement score by the average word-normalized cross-entropy from both models. Improbable sentence pairs will have higher average cross-entropy values. 

This score is also quite similar to the dual learning training criterion from \newcite{NIPS2016_6469} and \newcite{parity}. The dual learning criterion is formulated in terms of joint probabilities, later decomposed into translation model and language model probabilities. In practice, the influence of the language models is strongly scaled down which results in a form more similar to our score. 

While \citeauthor{moore-lewis:2010:Short} filtering requires an in-domain data set and a non-domain-specific data set to create helper models, we require a clean, relative high-quality parallel corpus to train the two dual translation models. We sample 1M sentences from WMT parallel data excluding Paracrawl and train Nematus-style translation models $W_{\mathrm{de}\rightarrow\mathrm{en}}$ and $W_{\mathrm{en}\rightarrow\mathrm{de}}$ (see Table~\ref{helper}).

Formula~(\ref{dualscore}) produces only positive values with 0 being the best possible score. We turn it into a partial score with values between 0 and 1 (1 being best) by negating and exponentiating, setting 
$A = W_{\mathrm{de}\rightarrow\mathrm{en}}$ and
$B = W_{\mathrm{en}\rightarrow\mathrm{de}}$:

\begin{dmath*}
\mathrm{adq}(x,y) = \exp(-( \left| H_{A}(y|x) - H_{B}(x|y) \right|\\ 
   + \frac{1}{2} \left( H_{A}(y|x) + H_{B}(x|y) \right))).
\end{dmath*}

Combining the $\mathrm{adq}$ filter with the $\mathrm{de\textrm{-}en}$ filter results in  a promising NMT system ($\mathrm{de\textrm{-}en}\cdot\mathrm{adq}$ in Table~\ref{dev}) trained on Paracrawl alone that beats the BLEU scores of the pure-WMT baseline. 

We further evaluated three ablative systems: 
\begin{itemize}
\item we omitted the language id filter (no de-en) which resulted in a system worse than randomly selected. This is not too surprising as we would expect many identical strings to be selected as highly adequate;
\item we dropped the absolute difference from formula (\ref{dualscore}) which decreased BLEU by about 1 point;
\item we removed the weighting by the averaged cross-entropies from formula (\ref{dualscore}), loosing about 3 BLEU points.
\end{itemize}
This seems to indicate that the two components of the dual conditional cross-entropy filter are indeed useful and that we have a practical scoring method for parallel data.

\subsection{Cross-entropy difference filtering}

When inspecting the training data generated with the above methods we saw many fragments that looked like noisy or not particularly useful data. This included concatenated lists of dates, series of punctuation marks or simply not well-formed text. Due to the adequacy filtering, the noise was at least adequate, i.e.~similar or identical on both sides and mostly correctly translated if applicable. The language filter had made sure that only few fully identical pairs of fragments had remained. 

However, we preferred to have a training corpus that also looked like clean data. To achieve this we treated
cross-entropy filtering proposed by \newcite{moore-lewis:2010:Short} as another score. Cross-entropy filtering or Moore-Lewis filtering uses the quantity
\begin{equation}
\begin{aligned}
H_{I}(x) - H_{N}(x)
\end{aligned}
\end{equation}
where $I$ is an in-domain model, $N$ is a non-domain-specific model and $H_M$ is the word-normalized cross-entropy of a probability distribution $P_M$ defined by a model $M$:
\begin{equation*}
\begin{aligned}
H_M(x) =& -\frac{1}{|x|}\log P_M(x) \\
 =& -\frac{1}{|x|} \sum_{t = 1}^{|x|}\log P_M(x_t|x_{<t}).
\end{aligned}
\end{equation*}
Sentences scored with this method and selected when their score is below a chosen threshold are likely to be more in-domain according to model $I$ and less similar to data used to train $N$ than sentences above that threshold. 

We chose WMT English news data from the years 2015-2017 as our in-domain, clean language model data and sampled 1M sentences to train model $I=W_{\mathrm{en}}$. We sampled 1M sentences from Paracrawl without any previously applied filtering to produce $N=P_{\mathrm{en}}$. The shared task organizers encourage submitting teams to not optimize for a specific domain, but it has been our experience that news data is quite general and clean data beats noisy data on many domains. 

To create a partial score for which the best sentence pairs produce a 1 and the worst at 0, we apply a number of transformations. First, we negate and exponentiate cross-entropy difference arriving at a quotient of perplexities of the target sentence $y$ ($x$ is ignored):
\begin{equation*}
\begin{aligned}
\mathrm{dom}^\prime(x,y) = & \; \exp(-(H_{I}(y) - H_{N}(y))) \\ = & \; \frac{\mathrm{PP}_{N}(y)}{\mathrm{PP}_{I}(y)}.
\end{aligned}
\end{equation*}
This score has the nice intuitive interpretation of how many times sentence $y$ is less perplexing to the in-domain model $W_{\mathrm{en}}$ than to the out-of-domain model $P_{\mathrm{en}}$. 

We further clip the maximum value of the score to 1 (the minimum value is already 0) as:
\begin{equation}
\begin{aligned}
\mathrm{dom}(x,y) = \min(\mathrm{dom}^\prime(x,y), 1).
\end{aligned}
\end{equation}
This seems counterintuitive at first, but is done to avoid that a high monolingual in-domain score strongly overrides bilingual adequacy; we are fine with low in-domain scores penalizing sentence pairs. This is a precision-recall trade-off for adequacy and we prefer precision.

Finally, we also propose a cut-off value $c$ as a parameter:
\begin{equation*}
\mathrm{cut}(x, c) = \left\{ \begin{array}{ll} x & \mathrm{if \;} x \geq c \\ 0 & \mathrm{otherwise} \\ \end{array} \right.
\end{equation*}
\begin{equation}
\mathrm{dom}_c(x,y) = \mathrm{cut}(\mathrm{dom}(x,y), c).
\end{equation}
Parameter $c$ can be used to completely eliminate sentence pairs, regardless of other scores, if $y$ is less than $c$ times more perplexing to the out-of-domain model than to the in-domain model, or inversely $1/c$ times more perplexing to the in-domain model than the out-of-domain model. This seems useful if we want a hard noise-filter similar to the language-id filter described above.

We used the domain filter only in combination with the previously introduced filters. In Table~\ref{dev}, we can observe that any variant leads to small improvements of the model over variants without the $\mathrm{dom}$ filters. This is expected as we optimized for WMT news development sets. We experimented with three cut-off values: 0.00 (no cut-off), 0.25 and 0.50, reaching the highest BLEU scores for a cut-off value $c=0.25$. This best result (bold in Table~\ref{dev}) was submitted to the shared task organizers as our only submission. 

Future work should consider bilingual cross-entropy difference filtering as proposed by \newcite{axelrod} where both sides of the corpus undergo the selection process or experiment with conditional probability distributions (translation models) for domain filtering.

\begin{table}[t]
\centering
\begin{tabular}{lcc}\toprule
Filter & test16 & test17 \\ \midrule
WMT18-full & 33.9 & 29.0\\ 
random & 16.2 & 14.1 \\ \midrule
de-en$\cdot$random & 26.6 & 23.3 \\ \midrule
de-en$\cdot$adq & 35.1 & 30.2 \\
\quad - no de-en & 15.4 & 12.7 \\
\quad - no absolute difference   & 33.8 & 29.3 \\
\quad - no CE weighting & 31.7 & 27.4 \\ \midrule
 de-en$\cdot$adq$\cdot$dom$_{0.00}$ & 35.5 & 30.5 \\
\bf de-en$\cdot$adq$\cdot$dom$_{\mathbf{0.25}}$ & \bf 36.0 & \bf 31.0 \\
de-en$\cdot$adq$\cdot$dom$_{0.50}$ & 35.4 & 30.6 \\ \midrule
de-en$\cdot$sim & 34.5 & 29.6 \\
de-en$\cdot$sim$\cdot$dom$_{0.25}$ & 35.5 & 30.6 \\
de-en$\cdot$adq$\cdot$sim$\cdot$dom$_{0.25}$ & 35.5 & 30.7 \\

\bottomrule
\end{tabular}
\caption{Results on development data. We only train neural models for the 100M sub-task. We did not optimize for any of the other three sub-tasks.}
\label{dev}
\end{table}

\subsection{Cosine similarity of sentence embeddings}
We further experimented with sentence embedding similarity to contrast this method with our cross-entropy based approach. 
Recently, \newcite{parity} and \newcite{schwenk} used cosine similarities of sentence embeddings in a common multi-lingual space to select translation pairs for neural machine translation. Both these approaches rely on creating a multi-lingual translation model across all available translation directions and then using the accumulated encoder representations (after summing or max-pooling contextual word-level embeddings across the time dimension) of sentences in a pair to compute similarity scores.

Following \newcite{parity}, we train a new multi-lingual translation model on WMT18 parallel data (excluding Paracrawl) by joining German-English and English-German training data into a mixed-direction training set (see model $W_{\mathrm{de}\leftrightarrow\mathrm{en}}$ in Table~\ref{helper}). 
For a given sentence $x$, we create its sentence embedding vector $\mathbf{s}_x$ according to  translation model $W_{\mathrm{de}\leftrightarrow\mathrm{en}}$ by collecting encoder representation vectors $\mathbf{h}_1$ to $\mathbf{h}_{|x|}$
 
 \begin{table*}[t]
\centering
\begin{subtable}[t]{7cm}\centering
\begin{tabular}{p{4cm}c} \toprule
System & Avg-BLEU \\ \midrule
RWTH Neural Redund. & 24.58 \\
RWTH Neural Indep. & 24.53 \\
\bf  Our submission  & \bf 24.45 \\
AliMT Mix  & 24.11 \\
AliMT Mix-div & 24.11 \\
\bottomrule
\end{tabular}
\caption{SMT 10M}
\end{subtable}\quad%
\begin{subtable}[t]{7cm}\centering
\begin{tabular}{p{4cm}c} \toprule
System & Avg-BLEU \\ \midrule
\bf  Our submission  & \bf 26.50 \\
AliMT Mix & 26.44  \\
AliMT Mix-div & 26.42 \\
Prompsit Active & 26.41 \\
NRC yisi-bicov & 26.40 \\
\bottomrule
\end{tabular}
\caption{SMT 100M}
\end{subtable}
\vspace{0.5cm}

\begin{subtable}[t]{7cm}\centering
\begin{tabular}{p{4cm}c} \toprule
System & Avg-BLEU \\ \midrule
\bf  Our submission  & \bf 28.62 \\
RWTH Neural Redund. & 28.01 \\
RWTH Neural Indep. & 28.00 \\
Speechmatics best & 27.97 \\
Speechmatics prime & 27.88 \\
\bottomrule
\end{tabular}
\caption{NMT 10M}
\end{subtable}\quad%
\begin{subtable}[t]{7cm}\centering
\begin{tabular}{p{4cm}c} \toprule
System & Avg-BLEU \\ \midrule
\bf Our submission  & \bf 32.05 \\
AliMT Mix & 31.93  \\
AliMT Mix-div & 31.92 \\
NRC yisi-bicov & 31.88 \\
NRC yisi & 31.76 \\
\bottomrule
\end{tabular}
\caption{NMT 100M}
\end{subtable}
\vspace{0.5cm}

\begin{subtable}[t]{9cm}\centering
\begin{tabular}{p{5cm}c} \toprule
System & Avg-BLEU \\ \midrule
\bf  Our submission  & \bf 111.63 \\
RWTH Neural Redundancy & 110.09 \\
AliMT Mix & 110.07  \\
AliMT Mix-div & 110.05 \\
RWTH Neural Independent & 109.91 \\
\bottomrule
\end{tabular}
\caption{Sum of all sub-tasks}
\end{subtable}

\caption{Top-5 out of 48 submissions for each of the four sub-tasks and total sum}\label{results}
\end{table*}

\begin{equation}
\mathbf{h}_{1:|x|} = \mathrm{Encoder}_{W_{\mathrm{de}\leftrightarrow\mathrm{en}}}(x)
\end{equation}
which are then averaged to form a single vector representation
\begin{equation}
\mathbf{s}_x = \frac{1}{|x|}\sum_{t=1}^{|x|} \mathbf{h}_t .
\end{equation}

For a given sentence pair $(x,y)$ we compute the cosine similarity of $\mathbf{s}_x$ and $\mathbf{s}_y$ as
\begin{equation}
\begin{aligned}
\mathrm{sim}(x,y) = \cos(\measuredangle\mathbf{s}_x \mathbf{s}_y) = \frac{\mathbf{s}_x \cdot \mathbf{s}_y}{|\mathbf{s}_x| |\mathbf{s}_y|}.
\end{aligned}
\end{equation}
Since the model has seen both languages, English and German, as source data it can produce useful sentence representations of both sentences in a translation pair. Unlike \newcite{parity}, we did not define a cut-off value for the similarity score as the threshold-based selection method of shared-task tool computes its own cut-off thresholds. 

We ran two experiments with the similarity based scores, evaluating configurations de-en$\cdot$sim and de-en$\cdot$adq$\cdot$sim$\cdot$dom$_{0.25}$. The first one corresponds to de-en$\cdot$adq and we compare the effectivness of the adq and sim filters after the application of the language-id-based filter de-en. We see in Table~\ref{dev} that while de-en$\cdot$sim leads to improvements over the language-filtered randomly selected Paracrawl data, it is significantly worse than de-en$\cdot$adq on both development sets. Interestingly, even when combined with our best scoring scheme (de-en$\cdot$adq$\cdot$dom$_{0.25}$) resulting in de-en$\cdot$adq$\cdot$sim$\cdot$dom$_{0.25}$ we see a slight degradation. Based on these results, we do not use the similarity scores for our submission.

In future experiments we want to use the multi-lingual model $W_{\mathrm{de}\leftrightarrow\mathrm{en}}$ instead of the two models $W_{\mathrm{en}\rightarrow\mathrm{de}}$ and $W_{\mathrm{de}\rightarrow\mathrm{en}}$ for our dual conditional cross-entropy method from Section~\ref{dual}. A multi-lingual model does not only have a common encoder, but also a common probability distribution for both languages which might lead to better agreement of the conditional cross-entropies.

\section{Shared task results}

As mentioned before, we submitted only our single-best set of scores $\mathrm{de\textrm{-}en}\cdot\mathrm{adq}\cdot\mathrm{dom}_{0.25}$ to the shared task. The shared task organizers trained four systems with each set of submitted scores, two Moses SMT \cite{conf/acl/KoehnHBCFBCSMZDBCH07} systems on the best 10M and 100M words corpora and two neural Marian NMT systems on the same sets. 

Based on the spread-sheet made available by the organizers, 48 sets of scores where submitted. Each set of scores was evaluated using the four mentioned models on 6 different test sets (newstest 2018, iwslt 2017, Acquis, EMEA, Global Voices, KDE). This required the organizers to train nearly 200 separate models; an effort that should be applauded. 

It seems that systems are ranked by their average score across these test sets and sub-tasks. In Table \ref{results} we selected the top-5 system across each sub-task for the purpose of this paper. The shared task overview will likely include a more thorough analysis. We place highest out of 48 submissions in three out of four tasks (SMT 100M, NMT 10M and NMT 100M) and third out of 48 for sub-task SMT 10M. The systems are packed quite closely, but the overall total across all four tasks shows, that we accumulate a slightly larger margin over the next best systems while the next four systems barely differ. This result is better than we expected as we only optimized for the NMT 100M task. 

For more details on the evaluation process and conclusions see the shared task overview paper \newcite{parallel-filtering-task:2018:WMT}.

\section{Future work and discussion}

We introduced dual conditional cross-entropy filtering for noisy parallel data and combined this filtering with multiple other noise filtering methods. Our submission to the WMT 2018 shared task on parallel corpus filtering achieved the highest overall rank and scored best in three out of four subtasks while scoring third in the fourth subtask. Each subtask had 48 participants. 

We believe this positive effect is rooted in the idea of directly asking a model that is very similar to the to-be-trained model which data it prefers (weighting by cross-entropy) while also constraining its answer with the introduced disagreement penalty. Our selection criterion is also very close to the optimization criterion used during NMT training, especially the dual learning training criterion. Other methods, for instance the evaluated similarity-based methods, do not have this direct connection to the training process. 

Future work should concentrate on further formalizing this method. We should analyze the connection to the dual learning training criterion on experiments whether models that were trained with this criterion are also better candidates for sentences scoring. Furthermore, the models we used for scoring were trained on small subsamples of clean data, we should investigate if stronger translation and language models are better discriminators. 



\bibliography{acl2018}
\bibliographystyle{acl_natbib_nourl}

\end{document}